# Sparse stochastic inference for latent Dirichlet allocation

**David Mimno**  MIMNO@CS.PRINCETON.EDU
Princeton U., Dept. of Computer Science, 35 Olden St., Princeton, NJ 08540, USA

**Matthew D. Hoffman**  MDHOFFMA@CS.PRINCETON.EDU
Columbia U., Dept. of Statistics, Room 1005 SSW, MC 4690 1255 Amsterdam Ave. New York, NY 10027

**David M. Blei**  BLEI@CS.PRINCETON.EDU
Princeton U., Dept. of Computer Science, 35 Olden St., Princeton, NJ 08540, USA

## Abstract

We present a hybrid algorithm for Bayesian topic models that combines the efficiency of sparse Gibbs sampling with the scalability of online stochastic inference. We used our algorithm to analyze a corpus of 1.2 million books (33 billion words) with thousands of topics. Our approach reduces the bias of variational inference and generalizes to many Bayesian hidden-variable models.

## 1. Introduction

Topic models are hierarchical Bayesian models of document collections (Blei et al., 2003). They can uncover the main themes that pervade a corpus and then use those themes to help organize, search, and explore the documents. In topic modeling, a "topic" is a distribution over a fixed vocabulary and each document exhibits the topics with different proportions. Both the topics and the topic proportions of documents are hidden variables. Inferring the conditional distribution of these variables given an observed set of documents is the central computational problem.

In this paper, we develop a posterior inference method for topic modeling that can find large numbers of topics in massive collections of documents. We demonstrate our approach by analyzing a collection of 1.2 million out-of-copyright books, comprising 33 billion observed words. Using our algorithm, we fit a topic model to this corpus with thousands of topics. We illustrate the most frequent words from several of these topics in Table 1.

Our algorithm builds on variational inference (Jordan et al., 1999). In variational inference, we define a parameterized family of distributions over the hidden structure—in this case topics and document-topic proportions—and then optimize the parameters to find a member of the family that is close to the posterior. Traditional variational inference for topic modeling uses coordinate ascent. The algorithm alternates between estimating document-topic proportions under the current settings of the topics and re-estimating the topics based on the estimated document proportions. This requires multiple passes through an entire collection, which is not practical when working with very large corpora.

Table 1. Randomly selected topics from a 2000-topic model trained on a library of 1.2 million out-of-copyright books.

| |
|---|
| killed wounded sword slain arms military rifle wounds loss |
| human Plato Socrates universe philosophical minds ethics |
| inflammation affected abdomen ulcer circulation heart |
| ships fleet sea shore Admiral vessels land boats admiral |
| sister child tears pleasure daughters loves wont sigh warm |
| sentence clause syllable singular examples clauses syllables |
| provinces princes nations imperial possessions invasion |
| women Quebec Women Iroquois husbands thirty whom |
| steam engines power piston boilers plant supplied chimney |
| lines points direction planes Lines scale sections extending |

Recently, Hoffman et al. (2010) introduced Online LDA, a stochastic gradient optimization algorithm for topic modeling. The algorithm repeatedly subsamples a small set of documents from the collection and then updates the topics from an analysis of the subsample. This method uses less memory than the standard approach because we do not need to store topic proportions for the full corpus. It also converges faster because we update topics more frequently. However, while it handles large corpora it does not scale to large numbers of topics.

Our algorithm builds on this method by using sampling to introduce a second source of stochasticity into the gradient. This approach lets us take advantage of sparse computation, scaling sublinearly with the number of topics. Using this algorithm, we can fit topic models to large collections with many topics.

## 2. Hybrid stochastic-MCMC inference

We model each of the $D$ documents in a corpus as a mixture of $K$ topics. This topic model can be divided into corpus-level *global* variables and document-level *local* variables. The global variables are $K$ topic-word distributions $\beta_1, ..., \beta_K$ over the $V$-dimensional vocabulary, each drawn from a Dirichlet prior with parameter $\eta$. For a document $d$ of length $N_d$, the local variables are (a) a distribution over topics $\theta_d$ drawn from a Dirichlet prior with parameter $\alpha$ and (b) $N_d$ token-topic indicator variables $z_{d1}, ..., z_{dN_d}$ drawn from $\theta_d$.

Our goal is to estimate the posterior distribution of the hidden variables given an observed corpus. We will use variational inference. Unlike standard mean-field variational inference, but similar to Griffiths & Steyvers (2004) and Teh et al. (2006), we will marginalize out the topic proportions $\theta_d$. Thus we need to approximate the posterior over the topic assignments $z_d$ and the topics $\beta_{1:K}$.

We will use a variational distribution of the form

$$q(z_1, ..., z_D, \beta_1, ..., \beta_K) = \prod_d q(z_d) \prod_k q(\beta_k). \quad (1)$$

This factorization differs from the usual mean-field family for topic models. Rather than defining a distribution that factorizes over individual tokens, we treat each document's sequence of topic indicator variables $z_d$ as a unit. As a result $q(z_d)$ will be a single distribution over the $K^{N_d}$ possible topic configurations, rather than a product of $N_d$ distributions, each over $K$ possible values.

We now derive an algorithm that uses Gibbs sampling to estimate variational expectations of the local variables and a stochastic natural gradient step to update the variational distribution of global variables. A lower bound on the marginal log probability of the observed words given the hyperparameters is

$$\log p(\boldsymbol{w}|\alpha, \eta) \geq \sum_d \mathbb{E}_q \log \left[ p(z_d|\alpha) \prod_i \beta_{z_{di} w_{di}} \right] \quad (2)$$
$$+ \sum_k \mathbb{E}_q \log p(\beta_k|\eta) + \mathcal{H}(q),$$

where $\mathcal{H}(q)$ denotes the entropy of $q$.

Following Bishop (2006), the optimal variational distribution over topic configurations for a document, holding all other variational distributions fixed, is

$$q^\star(z_d) \propto \exp\{\mathbb{E}_{q(\neg z_d)}[\log p(z_d|\alpha)p(w_d|z_d, \beta)]\} \quad (3)$$
$$= \frac{\Gamma(K\alpha)}{\Gamma(K\alpha + N_d)} \prod_k \frac{\Gamma(\alpha + \sum_i I_{z_{di}=k})}{\Gamma(\alpha)} \quad (4)$$
$$\times \prod_i \exp \mathbb{E}_q[\log \beta_{z_{di} w_{di}}]$$

where $I_{a=b}$ is 1 if $a = b$ and 0 otherwise, and $\neg z_d$ denotes the set of all unobserved variables besides $z_d$. We can compute Eq. 4 for a specific topic configuration $z_d$, but we cannot tractably normalize it to get the distribution $q^\star(z_d)$ over all $K^{N_d}$ configurations.

The optimal variational distribution over topic-word distributions, holding the other distributions fixed, is the kernel of a Dirichlet distribution with parameters

$$\lambda_{kw} = \eta + \sum_d \sum_i \mathbb{E}_q[I_{z_{di}=k} I_{w_{di}=w}]. \quad (5)$$

This expression includes the expectation under $q$ of the number of tokens of type $w$ assigned to topic $k$. Computing this expectation would require evaluating the intractable distribution $q^\star(z_d)$.

### 2.1. Online stochastic inference

We optimize the variational topic-word parameters $\lambda_{kw}$ using stochastic gradient ascent. Stochastic gradient ascent iteratively updates parameters with noisy estimates of the gradient. We obtain these noisy estimates by subsampling the data (Sato, 2001; Hoffman et al., 2010).

We first recast the variational objective in Eq. 2 as a summation over per-document terms $\ell_d$, so that the full gradient with respect to $\boldsymbol{\lambda}_k$ is the sum $\sum_d \frac{\partial}{\partial \boldsymbol{\lambda}_k} \ell_d$. We can then generate a noisy approximation to this full gradient by sampling a minibatch of documents $\mathcal{B}$ and then scaling the sum of the document-specific gradients to match the total size of the corpus,

$$\sum_d \frac{\partial}{\partial \boldsymbol{\lambda}_k} \ell_d = \mathbb{E} \left[ \frac{D}{|\mathcal{B}|} \sum_{d \in \mathcal{B}} \frac{\partial}{\partial \boldsymbol{\lambda}_k} \ell_d \right]. \quad (6)$$

(The expectation is with respect to the random sample $\mathcal{B}$.) Pushing the per-topic terms in Eq. 2 inside the summation over documents and removing terms not involving $\lambda_{kw}$ we obtain

$$\ell_d = \sum_w \left( \mathbb{E}_q[N_{dkw}] + \frac{1}{D}(\eta - \lambda_{kw}) \right) \mathbb{E}_q[\log \beta_{kw}] \quad (7)$$
$$+ \frac{1}{D} \left( \log \Gamma(\sum_w \lambda_{kw}) - \sum_w \log \Gamma(\lambda_{kw}) \right)$$

**Algorithm 1** Algorithm for hybrid stochastic variational-Gibbs inference.
   **for** $t \in 1, ..., \infty$ **do**
      $\rho_t \leftarrow \left(\frac{1}{t_0+t}\right)^\kappa$
      sample minibatch $\mathcal{B}$
      **for** $d \in \mathcal{B}$ **do**
         initialize $\boldsymbol{z}_d^0$
         discard $B$ burn-in sweeps
         **for** sample $s \in 1, ..., S$ **do**
            **for** token $i \in 1, ..., N_d$ **do**
               sample $z_{di}^s \propto (\alpha + N_{dk}) e^{\mathbb{E}_q[\log \beta_{kw}]}$
            **end for**
         **end for**
      **end for**
      $\lambda_{kw}^t \leftarrow (1-\rho_t)\lambda_{kw}^{t-1} + \rho_t\left(\eta + \frac{D}{|\mathcal{B}|}\hat{N}_{kw}\right)$
   **end for**

where $\mathbb{E}_q[N_{dkw}] = \sum_i \mathbb{E}_q[I_{z_{di}=k}I_{w_{di}=w}]$. The gradient of Eq. 7 with respect to the parameters $\lambda_{k1}, ..., \lambda_{kV}$ can be factored into the product of a matrix and a vector. The matrix, which contains derivatives of the digamma function, is the Fisher information matrix for the topic parameters. Element $w$ of the vector is

$$\mathbb{E}_q[N_{dkw}] + \frac{1}{D}(\eta - \lambda_{kw}). \quad (8)$$

Premultiplying the gradient of an objective function by the *inverse* Fisher information matrix of the distribution being optimized (in our case the variational distribution $q$) results in the natural gradient (Sato, 2001). Since our gradient is the product of the Fisher information matrix and a vector, the natural gradient is therefore simply Eq. 8 (Hoffman et al., 2010). Compared to the standard Euclidean gradient, the natural gradient offers both faster convergence (because it takes into account the information geometry of the variational distribution) and cheaper computation (because the vector in Eq. 8 is a simple linear function).

### 2.2. MCMC within stochastic inference

We cannot evaluate the expectation in Eq. 8 because we would have to consider a combinatorial number of topic configurations $\boldsymbol{z}_d$. To use stochastic gradient ascent, however, we only need an approximation to this expectation. We use Markov chain Monte Carlo to sample topic configurations from $q^\star(\boldsymbol{z}_d)$. We then use the empirical average of these samples to estimate the expectations needed for Eq. 8.

Gibbs sampling for a document starts with a random initialization of the topic indicator variables $\boldsymbol{z}_d$. We then iteratively resample the topic indicator at each position from the conditional distribution over that position given the remaining topic indicator variables:

$$q^\star(z_{di}=k|\boldsymbol{z}_{\setminus i}) \propto (\alpha + \sum_{j \neq i} I_{z_j=k})\exp\{\mathbb{E}_q[\log \beta_{kw_{di}}]\}, \quad (9)$$

where the expectation of the log probability of word $w$ given a topic $k$ is $\Psi(\lambda_{kw}) - \Psi(\sum_{w'} \lambda_{kw'})$. After $B$ burn-in sweeps, we begin saving sampled topic configurations. Once we have saved $S$ samples $\{\boldsymbol{z}\}^{1,...,S}$, we can define approximate sufficient statistics

$$\mathbb{E}_q[N_{dkw}] \approx \hat{N}_{kw} = \frac{1}{S}\sum_s\sum_{d \in \mathcal{B}}\sum_i I_{z_{di}^s=k}I_{w_{di}=w}. \quad (10)$$

Using MCMC estimates adds noise to our gradient, but allows us to use a collapsed objective function that does not represent document-topic proportions $\theta_d$. In addition, an average over a finite set of samples provides a sparse estimate of the gradient: for many words and topics, our estimate of $\mathbb{E}_q[N_{dkw}]$ will be zero.

### 2.3. Algorithm

We have defined a natural gradient and a method for approximating the sufficient statistics of that gradient. For a sequence of learning rates $\rho_t = (t_0 + t)^{-\kappa}$, the following update will lead to a stationary point:

$$\lambda_{kw}^t \leftarrow \lambda_{kw}^{t-1} + \rho_t \frac{D}{|\mathcal{B}|}\sum_{d \in \mathcal{B}}\left(\hat{N}_{dkw} + \frac{1}{D}(\eta - \lambda_{kw})\right)$$
$$= (1-\rho_t)\lambda_{kw}^{t-1} + \rho_t\left(\eta + \frac{D}{|\mathcal{B}|}\sum_{d \in \mathcal{B}}\hat{N}_{dkw}\right). \quad (11)$$

This update results in Algorithm 1. Two implementation details that result in sparse computations can be found in Appendix A. This online algorithm has the important advantage over Online LDA of preserving sparsity in the topic-word parameters, so that $\lambda_{kw} = \eta$ for most values of $k$ and $w$. Sparsity increases the efficiency of updates to $\boldsymbol{\lambda}_k$ and of Gibbs sampling for $\boldsymbol{z}_d$. Previous variational methods lead to dense updates to $KV$ topic parameters, making them expensive to apply to large vocabularies and large numbers of topics. Our method, in contrast, is able to exploit the sparsity exhibited by samples from the variational distribution $q^\star$, resulting in much more efficient updates.

## 3. Related Work

This paper combines two sources of zero-mean noise in constructing an approximate gradient for a variational inference algorithm: subsampling of data, and Monte Carlo inference. These sources of variance have been used individually in previous work. Stochastic approximation EM (SAEM, Delyon et al., 1999) combines an EM algorithm with a stochastic online inference procedure. SAEM does not subsample data, but rather interpolates between Monte Carlo estimates of the complete data. Kuhn & Lavielle (2004) extend SAEM to use MCMC estimates. Similarly, online EM (Cappé & Moulines, 2009) sub-samples data but preserves standard inference procedures for local variables.

Standard collapsed Gibbs sampling uses multiple sweeps over the entire corpus, representing topic-word distributions using the topic-word assignment variables of the entire corpus except for the current token. As a result, topic assignment variables must in theory be sampled sequentially, although parallel approximations work well empirically (Asuncion et al., 2008). In contrast, Algorithm 1 treats topic-word distributions as a global variable distinct from the local token-topic assignment variables, and so can parallelize trivially.

In this work we integrate over document-topic proportions $\theta_d$ within a variational algorithm. Collapsed variational inference (Teh et al., 2006) also analytically marginalizes over the topic proportions, but still maintains a fully factorized distribution over topic assignments at each position. The method described here does not restrict itself to such factored distributions, and therefore reduces bias, but this reduction may be offset by the bias we introduce when we initialize the Gibbs chain.

## 4. Empirical Results

In this section we compare the sampled online algorithm to two related online methods and measure the effect of model parameters. We use a selection of metrics to evaluate models.

### 4.1. Evaluation

**Held-out probability.** A model that characterizes the semantic structure of a corpus should place more of its probability mass on sensible documents than on random sequences of words. We can use this assumption to compare different models by asking each model to estimate the probability of a previously unseen document. A better model should, on average, assign higher probability to real documents than a lower-quality model. We evaluate held-out probability using the left-to-right sequential sampling method (Wallach et al., 2009; Buntine, 2009). For each trained model we generate point estimates of the topic-word probabilities $\tilde{p}(w|k)$. We then process each document by iterating through the tokens $w_1, ..., w_{N_d}$. At each position $i$ we calculate the marginal probability

$$\tilde{p}(w_i|w_{<i}) = \sum_k p(z_i = k|w_{<i}, z_{<i}, \alpha)\tilde{p}(w_i|k). \quad (12)$$

We then sample $z_i$ proportional to the terms of that summation and continue to the next token.[1] In order to normalize for document lengths, we divide the sum of the logs of the marginal probabilities by $N_d$.

**Coherence.** This metric measures the semantic quality of a topic by approximating the experience of a user viewing the $W$ most probable words for the topic (Mimno et al., 2011). It is related to point-wise mutual information (Newman et al., 2010). Let $D(w)$ be the document frequencies for each word $w$, that is, the number of documents containing one or more tokens of type $w$, and let $D(w_1, w_2)$ be the number of documents containing at least one token of $w_1$ and of $w_2$. For each pair of words $w_1, w_2$ in the top $W$ list, we calculate the number of documents that contain at least one token of the higher ranked word $w_1$ that also contain at least one token of the lower ranked word $w_2$:

$$\mathcal{C}(W) = \sum_i \sum_{j<i} \log \frac{D(w_i, w_j) + \epsilon}{D(w_j)} \quad (13)$$

where $\epsilon$ is a small constant used to avoid log zero. Values closer to zero indicate greater co-occurrence. Unlike held-out probability, which reports scores for held-out *documents*, coherence reports scores for individual *topics*.

**Wallclock time.** Our goal is to train useful models as efficiently as possible. In addition to model quality metrics, we are therefore also interested in total computation time.

### 4.2. Comparison to Online VB

Our first corpus consists of 350,000 research articles from three major journals: Science, Nature, and the Proceedings of the National Academy of Sciences of the USA. We use a vocabulary with 19,000 distinct words, including selected multi-word terms. We train models on 90% of the Science/Nature/PNAS corpus, holding out the remaining documents for testing purposes. We save topic-word parameters $\hat{N}_{kw}$ after epochs consisting of 500,000 documents.

Sampled online variational Bayesian inference compares well in terms of wallclock time to standard online VB inference, particularly with respect to the number of topics $K$. Figure 1 shows results comparing standard online VB inference to sampled online inference for $K$ up to 1000. Each iteration consists of a mini-batch of 100 documents. Standard online inference takes time linear in $K$, while wallclock time for sampled online inference grows more slowly.

---

[1] We set the document-topic hyperparameter $\alpha = 0.1$.

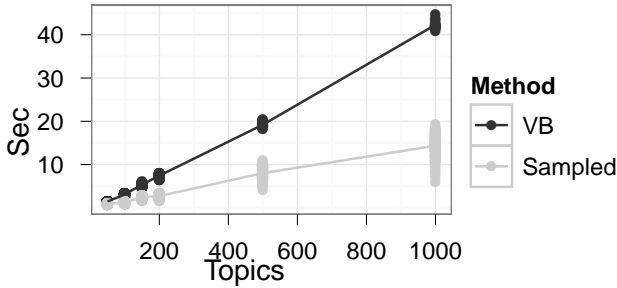

*Figure 1.* Comparison of seconds per mini-batch between online variational Bayes (Hoffman et al., 2010) and sampled online inference (this paper). Online VB is linear in $K$, while sampled inference takes advantage of sparsity.

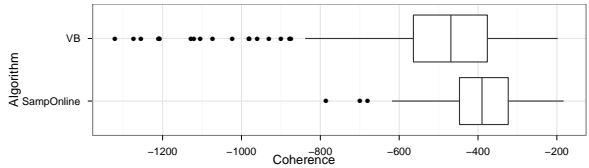

*Figure 2.* The new sampled online algorithm produces fewer low-quality topics than Online LDA at $K = 200$. Heldout log likelihood is much worse for Online LDA.

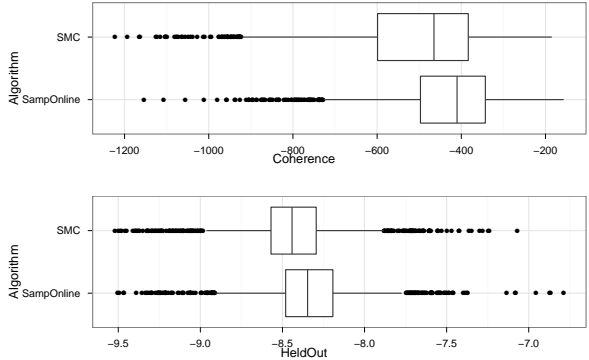

*Figure 3.* Sampled online inference performs better than one pass of sequential Monte Carlo, after processing a comparable number of documents with $K = 200$.

We would like to know if there is a difference in the quality of models trained through the hybrid sampled variational algorithm and the online LDA algorithm. We compare an implementation of Online LDA that tries to be as close as possible to the sampled online implementation, but using a dense VB update instead of a sparse sampled update for the local variables. In particular, the number of coordinate ascent steps in standard VB is equal to the number of Gibbs sweeps in the sampled algorithm.

Per-topic coherence for $K = 200$ is shown in Figure 2. Sampled online inference produces fewer very poor topics. This difference is significant under a two-sample t-test ($p < 0.001$) and does not decrease with additional training epochs. Sampled online inference also assigns greater held-out probability than Online LDA for every test document, by a wide margin. We evaluated several possible reasons for this difference in performance. Held-out probability estimation can be affected by evaluation-time smoothing parameter settings, but we found both models were affected equally. The log probability of a document is the sum of the log probabilities of its words. It is possible that if one model assigned very small probability to a handful of tokens, those words could significantly affect the overall score, but the difference in log probability was consistent across many tokens. The scale of parameters might not be comparable, but as both methods use the same learning schedule, the sum of the trained parameters $\lambda_{kw}$ is nearly identical.

The main difference appears to be the entropy of the topic distributions: the sampled-online algorithm produces less concentrated distributions (mean entropy $6.8 \pm 0.46$) than standard online LDA (mean entropy $6.0 \pm 0.58$). This result could indicate that coordinate ascent over the local variables for Online LDA is not converging.

### 4.3. Comparison to Sequential Monte Carlo

Sequential Monte Carlo is an online algorithm similar to Gibbs sampling in that it represents topics using sums over assignment variables (Ahmed et al., 2012). A Gibbs sampler starts with a random initialization for all hidden variables and sweeps repeatedly over the entire data set, updating each variable given the current value of all other variables. SMC samples values for hidden variables in sequential order, conditioning only on previously-seen variables. It is common to keep multiple sampling states or "particles", but this process adds both computation and significant bookkeeping complexity. Ahmed et al. (2012) use a single SMC state.

In order to compare SMC to the sampled online algorithm, we ran 10 independent SMC samplers over the Science/Nature/PNAS dataset, with documents ordered randomly. We also ran 10 independent sampled trainers, stopping after a number of documents had been sampled equivalent to the size of the corpus. In order to make the comparison more fair, we allowed the SMC sampler to sweep through each document the same number of times as the sampled online algorithm, but only the final topic configuration of a document was available to the subsequent documents.[2] Results for $K = 200$ are shown in Figure 3. SMC has consistently worse per-topic coherence and per-document held-out log probability. The sampled online algorithm in this paper differs from SMC in that the contribution of local token-topic assignment variables decays according to the learning rate schedule, so that more recently sampled documents can have greater weight than earlier documents. This decay allows sampled online inference to "forget" its initial topics, unlike SMC, which weights all documents equally.

### 4.4. Effect of parameter settings

**Number of samples.** In the inner loop of our algorithm we initialize[3] the topic indicator variables $z$ for a document and then perform several Gibbs sweeps. In each sweep we resample the value of each topic indicator variable in turn. We introduce bias when we initialize, so we discard $B$ "burn-in" sweeps and use values of $z$ saved after $S$ additional sweeps to calculate the gradient. Since performance is linear in the total number of sweeps $B + S$, we want to find the smallest number of sweeps that does not sacrifice performance.

We consider nine settings of the pair $(B, S)$. Under the first three settings we save one sweep and vary the number of burn-in sweeps: (1,1), (2,1), (3,1). For the second three settings we perform five sweeps, varying how many we discard: (2,3), (3,2), (4,1). The final three settings fix $B = S$ and consider larger total numbers of sweeps: (5,5), (10,10), (20,20). We evaluate each setting after processing 500,000 documents.

Performance was similar across settings with the following exceptions, which were significant at $p < 0.001$ under a two-sample $t$-test. The two-sweep setting (1,1) had better topic coherence but worse held-out probability than the all other settings. The (5,5) setting had the best mean held-out probability, but it was not significantly better than (10,10) and (20,20). The many-sweep settings (5,5), (10,10), (20,20) had worse topic coherence than the other settings, with many visibly low-quality topics. These results suggest that 3–5 sweeps is sufficient.

**Topic-word smoothing.** Eq. 9 involves the function $e^{\Psi(x)}$. This function approaches $x - \frac{1}{2}$ as $x$ gets large, but for values of $x$ near 0, it is non-linear. For example, $e^{\Psi(0.05)}$ is $10^{34}$ times greater than $e^{\Psi(0.01)}$. If the values of topic parameters are in this range, a minuscule increase in the parameter for word $w$ in topic $k$ can cause a profound change in the sampling distribution for that word: all subsequent tokens of type $w$ will be assigned to topic $k$ with probability near 1.0.

In general, the randomness introduced by sampling topic assignments helps to avoid becoming trapped in local maxima. When parameters are near zero, however, random decisions early in the inference process risk becoming permanent. The topic-word smoothing parameter $\eta$ can push parameter values away from this explosive region. We measured coherence for six settings of the topic-word hyperparameter $\eta$, $\{0.1, 0.2, 0.3, 0.4, 0.5, 0.6\}$. At $\eta = 0.1$, a common value for batch variational inference, many topics are visibly nonsensical. Average coherence improves significantly for each increasing value of $\eta \in \{0.2, 0.3, 0.4\}$ ($p < 0.001$). There is no significant difference in average coherence for $\eta \in \{0.4, 0.5, 0.6\}$.

**Forgetting factors.** We now consider the learning rate $\rho_t = (t_0 + t)^{-\kappa}$ and its relation to the corpus size $D$. We fix $\kappa = 0.6$ and vary the offset parameter $t_0 \in \{3000, 15000, 30000, 150000, 300000\}$, saving topic parameters after five training epochs of 500,000 documents each. There was no significant difference in average topic coherence.

The learning rate, however, is not the only factor that determines the magnitude of parameter updates. Eq. 11 also includes the size of the corpus $D$. If the corpus is larger, we will take larger steps, regardless of the contents of the mini-batch. The offset parameter $t_0$ had no significant effect on coherence for the full corpus, but it may have an effect if we also vary the corpus size.

We simulate different size corpora by subsampling the full data set. Results are shown in Figure 4 for models trained on one half, one quarter, and one eighth of the corpus. Each corpus is a subset of the next larger corpus. In the smallest corpus (12.5%), the model with $t_0 = 300000$ is significantly worse than other settings ($p < 0.001$). Otherwise, there is no significant difference in average topic coherence.

---

[2] Note that SMC has an advantage. In the sampled online algorithm we Gibbs sample each document within a mini-batch independently, while in SMC, documents "see" results from *all* previous documents.

[3] We initialize by sampling each token conditioned on the topics of the previous tokens in the document: $p(z_{di} = k) \propto (\alpha + \sum_{j<i} I_{z_{dj}=k})p(w_{di}|k)$.

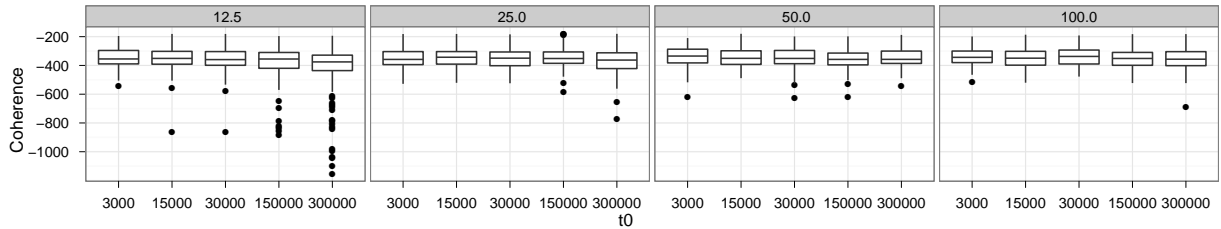

Figure 4. Topic quality is lowest for large values of $t_0$, but only in small corpora. Panels represent the proportion of training data used. Each panel shows coherence values for five $K = 100$ topic models with varying learning rates.

### 4.5. Scalability

**Pre-1922 books.** To demonstrate the scalability of the method, we modeled a collection of 1.2 million out-of-copyright books. Topic models are useful in characterizing the contents of the corpus and supporting browsing applications: even scanning titles for a collection of this size is impossible for one person. Previous approaches to million-book digital libraries have focused on keyword search and word frequency histograms (Michel et al., 2011). Such methods do not account for variability in meaning or context. There is no guarantee that the words being counted match the meaning assumed by the user. In contrast, an interface based on a topic model could, for example, distinguish uses of the word "strain" in immunology, mechanical engineering, and cookery.

We divide each book into 10-page sections, resulting in 44 million "documents" with a vocabulary size of $2^{16}$. We trained models with $K \in \{100, 500, 1000, 2000\}$. Randomly selected example topics are shown in Table 1, illustrating the average level of topic quality. Models are sparse: at $K = 2000$, less than 1% of the $2000 \cdot 2^{16}$ possible topic-word parameters are non-zero. The algorithm scales well as $K$ increases. The number of milliseconds taken to process a sequence of 10,000 documents was similar for $K = 1000$ and $2000$, despite doubling the number of topics.

## 5. Conclusions

Stochastic online inference allows us to scale topic modeling to large document sets. Sparse Gibbs sampling allows us to scale to large numbers of topics. The algorithm presented in this paper combines the advantages of these two methods. As a result, models can be trained on vast, open-ended corpora without requiring access to vast computer clusters. If parallel architectures are available, we can trivially parallelize computation within each mini-batch. As this work is related to the Online LDA algorithm of Hoffman et al. (2010), extensions to that model are also applicable, such as adaptive scheduling algorithms (Wahabzada & Kersting, 2011). The use of MCMC within stochastic variational inference reduces one source of bias in estimating local variables. Although we have focused on text analysis applications, this hybrid method generalizes to a broad class of Bayesian models.

## Acknowledgments

John Langford, Iain Murray, Charles Sutton provided helpful comments. Yahoo! and PICSciE provided computational resources. DM is supported by a CRA CI fellowship. MDH is supported by NSF ATM-0934516, DOE DE-SC0002099, and IES R305D100017. DMB is supported by ONR N00014-11-1-0651, NSF CAREER 0745520, AFOSR FA9550-09-1-0668, the Alfred P. Sloan foundation, and a grant from Google.

Sparse stochastic inference for latent Dirichlet allocation

## A. Sparse computation

**Sparse sampling over topics.** Sampling $z_{di}^s \propto (\alpha + N_{dk})e^{\mathbb{E}_q[\log \beta_{kw}]}$ requires calculating the normalizing constant $\mathcal{Z} = \sum_k (\alpha + N_{dk})e^{\mathbb{E}_q[\log \beta_{kw}]}$. This calculation can be accomplished in time much less than $O(k)$ if we can represent the topic-word parameters $\lambda_{kw}$ sparsely. The smoothing parameter $\eta$ can be factored out of Equation 11 as long as we assume that all initial values $\lambda_{kw}^0 \geq \eta$. Rearranging this equation to separate the Dirichlet hyperparameter $\eta$

$$\lambda_{kw}^t \leftarrow \eta + (1-\rho_t)\left(\lambda_{kw}^{t-1} - \eta\right) + \rho_t \frac{D}{|\mathcal{B}|} N_{kw}^{\mathcal{S}} \quad (14)$$

shows that we can define an alternative parameter $\tilde{N}_{kw}^t = \lambda_{kw}^t - \eta$ that represents the "non-smoothing" portion of the variational Dirichlet parameter, and ignore the contribution of the smoothing parameter until it is time to calculate expectations.

For any given $w$, it is likely that most values of $\tilde{N}_{kw}$ will be zero. We can therefore rewrite the normalizing constant as

$$\mathcal{Z} = \sum_k \frac{\alpha + N_{dk}}{e^{\Psi(V\eta + \tilde{N}_{k\circ})}} \left(e^{\Psi(\eta + \tilde{N}_{kw})} - e^{\Psi(\eta)}\right) +$$
$$\sum_k \frac{\alpha + N_{dk}}{e^{\Psi(V\eta + \tilde{N}_{k\circ})}} e^{\Psi(\eta)}. \quad (15)$$

The second summation does not depend on any word-specific variables, and can therefore be calculated and then updated incrementally as $N_{dk}$ changes. The first summation is non-zero only for $k$ such that $\tilde{N}_{kw} > 0$.

**Sparse updates in the vocabulary.** We expect that a typical mini-batch will contain a small fraction of the words in the vocabulary. Eq. 11, however, updates $\tilde{N}_{kw}$ for all words, even words that do not occur in the current mini-batch. Expanding the recursive definition of $\tilde{N}_{kw}^t$, and letting $\hat{N}_{kw}^t = \frac{D}{|\mathcal{B}|} N_{kw}^{\mathcal{S}}$,

$$\tilde{N}_{kw}^t = \rho_t \hat{N}_{kw}^t + (1-\rho_t)\left(\rho_{t-1}\hat{N}_{kw}^{t-1} + (1-\rho_{t-1})(\ldots)\right)$$
$$(16)$$

$$= \rho_t \hat{N}_{kw}^t + (1-\rho_t)\rho_{t-1}\hat{N}_{kw}^{t-1} + (1-\rho_t)(1-\rho_{t-1})\ldots$$
$$(17)$$

Dividing both sides by $\prod_{i=1}^t (1-\rho_i)$,

$$\frac{\tilde{N}_{kw}^t}{\prod_{i=1}^t (1-\rho_i)} = \frac{\rho_t \hat{N}_{kw}^t}{\prod_{i=1}^t (1-\rho_i)} + \frac{\rho_{t-1} \hat{N}_{kw}^{t-1}}{\prod_{i=1}^{t-1}(1-\rho_i)} \quad (18)$$
$$+ \frac{\rho_{t-2} \hat{N}_{kw}^{t-2}}{\prod_{i=1}^{t-2}(1-\rho_i)} + \ldots.$$

Defining a variable $\pi_t = \prod_{i=1}^t (1-\rho_i)$, the update becomes

$$\frac{\tilde{N}_{kw}^t}{\pi_t} = \frac{\tilde{N}_{kw}^{t-1}}{\pi_{t-1}} + \frac{\rho_t \hat{N}_{kw}^t}{\pi_t}. \quad (19)$$

This update is sparse: only elements with non-zero $n_{dw}$ will be modified. To calculate the expectation of $p(w|k)$, we compute $\Psi\left(\eta + \pi_t \frac{N_{kw}^t}{\pi_t}\right) - \Psi\left(W\eta + \pi_t \sum_w \frac{N_{kw}^t}{\pi_t}\right)$.

The scale factor $\pi_t$ can become small after several hundred mini-batches. We periodically "reset" this parameter by setting all stored values to $\tilde{N}_{kw}^t = \pi_t \frac{\tilde{N}_{kw}^t}{\pi_t}$, avoiding the possibility of numerical instability.